%% file: force_copy_workshop.tex
\pdfoutput=1
\documentclass[11pt]{article}

\usepackage[]{emnlp2021}

\usepackage{times}
\usepackage{latexsym}
\usepackage{enumitem}
\usepackage[T1]{fontenc}

\usepackage[utf8]{inputenc}

\usepackage{microtype}
\usepackage{graphicx}
\usepackage{array}
\usepackage{booktabs}
\usepackage{adjustbox}
\usepackage{multirow}
\newcolumntype{M}[1]{>{\arraybackslash}p{#1}}
\newcolumntype{L}[1]{>{\raggedright\arraybackslash}p{#1}}
\newcolumntype{C}[1]{>{\centering\arraybackslash}p{#1}}
\newcolumntype{R}[1]{>{\raggedleft\arraybackslash}p{#1}}
\definecolor{theblue}{RGB}{0, 112, 192}
\newcommand\rotowire{{\normalsize R\small OTO\normalsize W\small IRE\normalsize{ }}}
\newcommand\rotowirestrip{{\normalsize R\small OTO\normalsize W\small IRE\normalsize{}}}

\title{May the Force Be with Your Copy Mechanism: Enhanced Supervised-Copy Method for Natural Language Generation}

\author{Sanghyuk Choi, \ Jeong-in Hwang, \ Hyungjong Noh, \ Yeonsoo Lee \\
        NCSOFT NLP Center \\
        \texttt{\{sanghyuk, jihwang, nohhj0209, yeonsoo\}@ncsoft.com}}

\begin{document}
\maketitle
\begin{abstract}
Recent neural sequence-to-sequence models with a copy mechanism have achieved remarkable progress in various text generation tasks. These models addressed out-of-vocabulary problems and facilitated the generation of rare words. However, the identification of the word which needs to be copied is difficult, as observed by prior copy models, which suffer from incorrect generation and lacking abstractness. In this paper, we propose a novel supervised approach of a copy network that helps the model decide which words need to be copied and which need to be generated. Specifically, we re-define the objective function, which leverages source sequences and target vocabularies as guidance for copying. The experimental results on data-to-text generation and abstractive summarization tasks verify that our approach enhances the copying quality and improves the degree of abstractness.
\end{abstract}

\section{Introduction}

Natural language generation is a task in any setting in which we generate new text, such as machine translation, summarization, and dialogue systems \citep{gatt2018survey}. Approaches using neural networks known as sequence-to-sequence (seq2seq) have achieved reliable results on such tasks \citep{sutskever2014sequence}. However, basic seq2seq models exhibit weaknesses in dealing with rare and out-of-vocabulary (OOV) words. \citet{vinyals2015pointer} resolved this problem by copying words from the source sequence and directly inserting them into the output generation. This idea has been successfully applied to abstractive summarization tasks \citep[]{gulcehre-etal-2016-pointing, gu-etal-2016-incorporating, see-etal-2017-get, xu-etal-2020-self} and has been further adopted by \citet{wiseman-etal-2017-challenges} and \citet{puduppully2019data} to help the model copy correct values in data-to-text generation tasks.
\begin{table}[t]
\small
\centering
\begin{tabular}{C{1cm}C{0.6cm}C{0.7cm}C{0.6cm}C{0.95cm}C{0.55cm}C{0.2cm}}%
\toprule
TEAM   & WIN & LOSS & PTS & FG\_PCT & RB & ... \\ \midrule
Hawks  & 28  & 20   & 142 & 44      & 64 & ... \\
Knicks & 21  & 28   & 139 & 40      & 63 & ... \\ \addlinespace[\aboverulesep] \toprule 
\end{tabular}
\begin{tabular}{C{2cm}C{0.45cm}C{0.45cm}C{0.45cm}C{0.45cm}C{0.6cm}C{0.2cm}}
PLAYER          & AS  & RB  & PT  & FG  & FGA & ... \\ \midrule
Paul Millsap  & 7   & 19  & 37  & 13  & 29  & ... \\
Kent Bazemore & 0   & 10  & 37  & 13  & 29  & ... \\
...           & ... & ... & ... & ... & ... & ... \\ \bottomrule
\end{tabular}
\begin{tabular}{c}
\\
\end{tabular}
\setlength\extrarowheight{2pt}
\begin{tabular}{|M{7.3cm}|}
\hline
\textbf{Pointer-Generator} The \textcolor{theblue}{Atlanta Hawks defeated} \textcolor{red}{the host} \textcolor{theblue}{New York Knicks}, \textcolor{theblue}{142-139}, \textcolor{theblue}{at Philips Arena} on Wednesday. The Hawks came into this game as a sizable favorite and they didn't disappoint. In fact, \textcolor{red}{Atlanta led for over 40 minutes} of this game, \textcolor{red}{as they led by double - digits for the entirety of the second half.} ...\\ \hline
\textbf{Our Force-copy-unk} \textcolor{theblue}{The Atlanta Hawks} (\textcolor{theblue}{28-20}) \textcolor{theblue}{defeated} the \textcolor{theblue}{New York Knicks} (\textcolor{theblue}{21-28}) \textcolor{theblue}{142-139} at \textcolor{theblue}{Phillips Arena} in Atlanta. The \textcolor{theblue}{Hawks} were led by \textcolor{theblue}{Paul Millsap}, who \textcolor{theblue}{scored 37 points} (\textcolor{theblue}{13-29 FG, 3-8 3Pt, 8-10 FT}) to go with \textcolor{theblue}{19 rebounds, seven assists and one steal} in \textcolor{theblue}{60 minutes} ...\\\hline
\end{tabular}
\captionof{figure}{Comparison of output of data-to-text models on a \rotowire dataset. Text that accurately reflects a record is highlighted in \textcolor{theblue}{blue}, and erroneous text is highlighted in \textcolor{red}{red}.}
\label{figure1}
\end{table}

Unfortunately, the generation results obtained from copying mechanisms can be suboptimal. \citet{zhou2018sequential} observed that some unrelated words appear unexpectedly in the middle of the phrase, or the phrase is not copied completely and some words are missing. \citet{see-etal-2017-get} and \citet{gehrmann-etal-2018-bottom} reported the lack of abstractness in the excessive copying of source words, resulting in insufficient novel expressions of generated summarization. For example, the \emph{CNN/DailyMail} dataset \citep{nallapati-etal-2016-abstractive} contains approximately 17\% of novel words on gold summaries, but most copy models create less than 1\%. Moreover, most data-to-text models based on copy mechanisms still suffer from poor factual inconsistencies. Accurately conveying the facts is especially important in informational communication, such as news, and low levels of veracity make these models unreliable and useless in practice \citep{kryscinski-etal-2020-evaluating}. \citet{wiseman-etal-2017-challenges} indicated that there is a significant gap between neural models and template systems in terms of generating text containing factual (i.e., correct) records. Despite the efforts of the dedicated research community \citep[]{puduppully2019data, rebuffel2020hierarchical, rebuffel2021controlling} there are still many challenges to narrow the gap. 

In our experiments, we found that the probability of the copying words crucially affects the decoding process, resulting in fallacious generation when calculated inaccurately. Therefore, we argue that deciding between copying or generating can benefit from distinct guiding. Accordingly, we propose a \emph{force-copy} method that forces the model to copy every word when it appears in source and target sequences simultaneously so as to increase copy precision. Furthermore, to relieve the lacking abstractness of the generated text, we present the \emph{force-copy-unk} method, which forces model to copy a word only if it does not exist in the target vocabulary even if it appears in both the source and target sequences.

Our contributions are three-fold:
\begin{itemize}
  \item We analyzed and compared the characteristics of prior copy models. On the basis of the analysis, we present \emph{force-copy} and \emph{force-copy-unk} methods that promote accurate copy of source sequence. %
  \item The \emph{force-copy} and \emph{force-copy-unk} methods improve the RG Precision score in data-to-text, which indicates that the text generation based on the source data is conducted more correctly.
  \item The \emph{force-copy-unk} achieves an improvement in copy precision, which not only increases the ROUGE score but also shows better abstraction in abstractive summarization task.
\end{itemize}

\section{Background}
As our work builds on the prior copy mechanism, we introduce Pointer-Generator Network \citep{see-etal-2017-get} as a baseline. In this study, the source text \(x\) is fed into a bidirectional LSTM (BiLSTM) encoder. However \citet{klein-etal-2020-opennmt} reported that there is no significant difference in the copying performances of Transformer \citep{vaswani2017attention} encoder compared to that of BiLSTM; thus we replaced it with a Transformer encoder producing a sequence of encoded hidden states \(h_i\). At each timestep \(t\), the decoder receives the representation of the previously generated word to produce the decoder hidden states \(s_t\). From the hidden states, a context vector \(c_t\) is calculated based on the attention distribution \citep{bahdanau2015neural} as follows:
\begin{equation}
e_{t,i} = v^T\mathrm{tanh}(W_h h_i + W_s s_t)
\end{equation}\begin{equation}
\alpha_t = \mathrm{softmax}(e_t)
\end{equation}\begin{equation}
c_t = \sum\mathop{}_{\mkern-5mu i}\alpha_{t,i} h_i
\end{equation}
where \(v\), \(W_h\) and \(W_s\) are learnable parameters.

The vocabulary distribution \(P_{vocab}\) over all words in the target vocabulary is computed from \(s_t\), \(c_t\), and the learnable parameters \(W_{v}^{'}\), \(W_v\), \(b\) and \(b'\):
\begin{equation}
P_{vocab}=\mathrm{softmax}(W'_v (W_v [s_t;c_t]+b)+b')
\end{equation}

The \emph{generation probability} \(p_{gen}\) is used as a soft switch to either generate from vocabulary or copy from the source words by sampling from the attention distribution \(\alpha_t\). Additionally, \(p_{gen}\) for timestep \(t\) is calculated from the context vector \(c_t\), decoder state \(s_t\), and decoder input \(y_t\):
\begin{equation}
p_{gen}=\mathrm{sigmoid}(w_h^T c_t+w_s^T s_t+w_y^T y_t+b_{ptr})
\end{equation}
where vectors \(w_h^T\), \(w_s^T\), \(w_y^T\) and the scalar bias \(b_{ptr}\) are learnable parameters. Let a word \(w\) be an element in the set of extended vocabulary, which refers to the union of the target vocabulary and all words appearing in the input sequence; then, the final probability distribution \(P(w)\) is computed as:
\begin{equation}\label{final_distribution}
P(w)=p_{gen}P_{vocab}(w)+(1-p_{gen})\textstyle{\sum_{i:w_i=w}\alpha_{t,i}}
\end{equation}
If \(w\) is an OOV word, then \(P_{vocab}(w)\) is zero. Similarly, if \(w\) does not exist in the input sequence, then $\sum_{i:w_i=w}\alpha_{t,i}$ is zero. 
During training, the loss at timestep \(t\) is the negative log-likelihood of the target word \(w_t^*\) for that timestep:
\begin{equation}
loss^t=-\mathrm{log}P(w_t^*)
\end{equation}

\section{Model}
\begin{figure*}
\centering
\includegraphics[width=\textwidth]{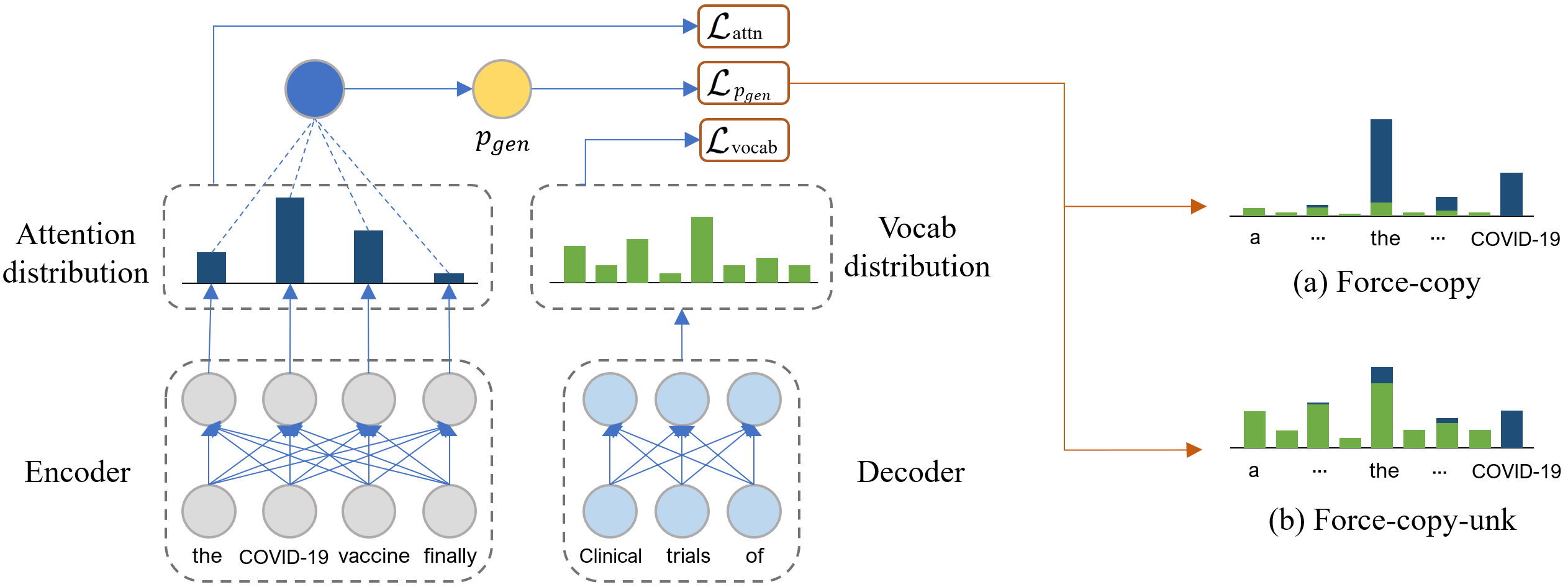}
\caption{\label{citation-guide}
Architecture of our models. A generation probability \(p_{gen}\) at each decoder timestep is trained through explicit loss function. (a) Force-copy model trains to copy if the target word exists in the source context. On the other hands, (b) Force-copy-unk model trains to generate from the vocab distribution unless the target word is an OOV. Note that \emph{COVID-19} is an out-of-vocabulary in this example. Best viewed in color.
}
\label{figure2}
\end{figure*}
We next consider techniques for incorporating supervised learning of the copying decision into the model. To identify the number of occasions at which every copy and generation occurs, Eq. (\ref{final_distribution}) is split into three cases: (i) No word is copied from the source sequence, i.e., words are sampled from the vocabulary probability distribution. In this case, \(p_{gen}\) is guided to take higher value during the 
training time; (ii) The word copied from the source sequence does not exist in the target vocabulary. In this case, \(p_{gen}\) is guided to take lower value during the training time; (iii) The word copied from the source sequence exists in the target vocabulary. In this case, \(p_{gen}\) could have a scalar range in \([0, 1]\) and is learned implicitly.

We augment the decoder by supervising a soft switch \(p_{gen}\) that determines whether the model generates or copies. First, we re-define the loss function as
\begin{equation}
loss^{t} = loss_{vocab}^{t} + loss_{attn}^{t} + loss_{p_{gen}}^{t}
\end{equation}
where \(loss_{vocab}^{t}\) is the maximum likelihood estimation (MLE), which is used as a standard training criterion in the sequence-to-sequence model:
\begin{equation}\label{loss_vocab}
loss_{vocab}^{t} = -\mathrm{log}(P_{vocab}(w_t^*))
\end{equation}
It should be noted that \(P_{vocab}(w_t^*)\) indicates the probability of generating the unknown token if \(w_t^*\) is OOV.
Similarly to the pointer network \citep{vinyals2015pointer}, we use attention distribution as a guide for copying, and train it to minimize the negative log-likelihood:
\begin{equation}\label{loss_attn}
loss_{attn}^{t} = -\mathrm{log}(\textstyle{\sum_{i:w_i=w}\alpha_{t,i}})
\end{equation}
Finally, we adopt \(loss_{p_{gen}}^t\) to train \(p_{gen}\) explicitly. In fact, the optimal way is to feed the gold copy answer for every target word; However, there rarely exist supervised data for this task. Hence, we suggest two methods leveraging the source sequences and target vocabularies.
\begin{table*}
\centering
\small
\begin{tabular}{@{}L{3.85cm}|L{1.25cm}|L{1.3cm}|L{1.4cm}|L{1.3cm}|L{1.3cm}|L{1.2cm}|L{1.32cm}@{}}
\toprule
Models & \begin{tabular}[c]{@{}l@{}}Use\\ explcit\\ switch\end{tabular} & \begin{tabular}[c]{@{}l@{}}Recycle\\ attention\end{tabular} & \begin{tabular}[c]{@{}l@{}}Conditional\\ activate\end{tabular} & \begin{tabular}[c]{@{}l@{}}Mix copy,\\ gen prob\end{tabular} & \begin{tabular}[c]{@{}l@{}}Sum src\\ duplicatd\\ word\end{tabular} & \begin{tabular}[c]{@{}l@{}}Force\\ train\\ switch\end{tabular} & \begin{tabular}[c]{@{}l@{}}Tradeoff\\ between\\ copy/gen\end{tabular}\\ \midrule
\citet{miao-blunsom-2016-language} & Yes & Yes & No & Yes & No & No & Yes\\
COPYNET \citep{gu-etal-2016-incorporating} & No & No & No & No & Yes & - & Yes\\ %
\citet{merity2016pointer} & No & Yes & No & Yes & Yes & - & Yes\\ %
\citet{gulcehre-etal-2016-pointing} & Yes & Yes & Yes & No & No & Yes & Yes\\ %
\citet{nallapati-etal-2016-abstractive} & Yes & Yes & Yes & No & No & Yes & Yes\\ %
PGNet \citep{see-etal-2017-get} & Yes & Yes & No & Yes & Yes & No & Yes\\ %
\citet{chen-etal-2020-shot} & Yes & Yes & Yes & Yes & Yes & Yes & Yes\\ %
\citet{wu-etal-2020-question} & Yes & Yes & Yes & Yes & Yes & Yes & Yes\\ %
OpenNMT + force copy attn & Yes & Optional & No & Yes & Yes & Yes & Yes\\ %
Force-copy (ours) & Yes & Yes & No & Yes & Yes & Yes & No\\ %
Force-copy-unk (ours) & Yes & Yes & No(partly) & Yes & Yes & Yes & No\\ \bottomrule
\end{tabular}
\caption{\label{citation-guide}
Brief comparison between prior copy models. \emph{Use explicit switch} denotes that the model calculates an explicit switch probability. \emph{Recycle attention} denotes that the model recycles the attention distribution as the copy distribution. \emph{Conditional activate} denotes that the model activates the copy mechanism only for unknown or named-entity words or keywords. \emph{Mix copy, gen prob} denotes that the model combines the probabilities from the vocabulary distribution with probabilities from the copy distribution. \emph{Sum src duplicated word} denotes that the model adds all probabilities of the same words from the attention distribution when a word appears multiple times in the source sequence. \emph{Force train switch} denotes that the model trains the switch probability explicitly. \emph{Tradeoff between copy/gen} denotes that the model has a loss trade-off between copy and generation during training.
}
\label{table1-copymodels}
\end{table*}
\vspace{5pt}\\
\textbf{Force-copy} Given a source sequence $ X = (x_1,x_2,...,x_{T_{x}})$, if a target word \(w_t^*\) appears in \(X\), \(w_t^*\) can be a \emph{copy-candidate}. For example, the word \emph{'the'} and \emph{'COVID-19'} are both copy-candidates in Figure \ref{figure2}. In the force-copy model, we assume that every copy-candidate word is copied from the source sequence. Therefore, \(p_{gen}\) can be derived from the loss function:
\begin{equation}
loss_{p_{gen}}^t =
  \left\{\begin{array}{l@{}l@{\qquad}l}
    -\mathrm{log}(1-p_{gen})\;if\;w_{t}^{*} \in X \\[\jot]
    -\mathrm{log}(p_{gen})\;otherwise
  \end{array}\right.
\end{equation}
This forces the copy switch \(p_{gen}\) to perform a copy all copy-candidate words. However, we do not penalize the generating ability even on copy circumstances. We always train \(loss_{vocab}^t\) on the loss function regardless of the copy process.
\vspace{5pt}\\
\textbf{Force-copy-unk} Whereas the force-copy model tries to copy every copy-candidate word, it is also possible to generate words from the vocabulary distribution instead of copying if the word exists in the target vocabulary. For example, the copy-candidate word \emph{'the'} could be generated instead of being copied in Figure \ref{figure2}. In this way, we restrict the scope of copy to unknown words in the force-copy-unk model. Therefore, \(p_{gen}\) can be derived from the loss function:
\begin{equation}\label{eq:force-copy-unk}
loss_{p_{gen}}^{t} =
  \left\{\begin{array}{l@{}l@{\qquad}l}
    -\mathrm{log}(1-p_{gen})\;if\;w_{t}^{*} \in X,\notin V\\[\jot]
    -\mathrm{log}(p_{gen})\;otherwise
  \end{array}\right.
\end{equation}
where \(V\) is a set of target vocabularies. This loss function forces the copy switch \(p_{gen}\) to copy the word only if it is an unknown token. For other copy-candidate words that are not OOV, we found that it is effective to utilize the copy information as well (see Section \ref{sec:effect-of-cp}). Inspired by work in guided alignment training for machine translation \citep{chen2016guided}, we retain the \(loss_{attn}^{t}\) on the loss function to inform the decoder that the word is a copy-candidate, thus inducing \emph{copy-like} generation.

\section{Related Work}
\textbf{Abstractive Summarization}
\\Abstractive summarization aims to generate accurate and concise summaries that contain novel words in contrast to extractive summarization that extracts almost entire sentences from the input. Most prior works on abstractive summarization that employed neural networks achieved inspiring results \citep[]{rush-etal-2015-neural, nallapati-etal-2016-abstractive, see-etal-2017-get}. Especially after \citet{vinyals2015pointer}, \citet{gulcehre-etal-2016-pointing} showed dramatic improvement by applying copy mechanisms, it has become a primary module for abstractive summarization. Recent work leveraging such pre-training models 
\citep[]{NEURIPS2019_c20bb2d9, song2019mass, lewis-etal-2020-bart, zhang2020pegasus} has presented impressive advancements in performance when fine-tuned for text generation tasks. Applying the copy mechanism to these pre-training models has extended the success \citep[]{xu-etal-2020-self, bi-etal-2020-palm}.
\\\textbf{Data-to-text Generation}
\\Data-to-text generation tasks aim to produce texts from non-linguistic input data \citep{reiter-2007-architecture}, including the generation of weather forecasts \citep{liang-etal-2009-learning} and biographical content from Wikipedia \citep{lebret-etal-2016-neural}. Specifically, \citet{wiseman-etal-2017-challenges} showed that a neural encoder-decoder model powered with the copy mechanism \citep[]{gu-etal-2016-incorporating, gulcehre-etal-2016-pointing} can generate fluent multi-sentence summaries from game statistics without explicit templates or rules. Based on their approach, various promising neural approaches have been proposed for \rotowire tasks \citep[]{puduppully2019data, puduppully-etal-2019-data, rebuffel2020hierarchical, iso-etal-2019-learning}. Regarding the copy mechanism, \citet[]{puduppully2019data, puduppully-etal-2019-data} and \citet{iso-etal-2019-learning} adopted the same method as \citet{wiseman-etal-2017-challenges}, and \citet{rebuffel2020hierarchical} utilized Pointer-Generator Network (PGNet) \citep{see-etal-2017-get}\footnote{We analyzed released code (if available) to identify which copy mechanism they employed in case it was not mentioned in their paper.}.
\\\textbf{Pointer / Copy Mechanism}
\\The fundamental structure of the copy mechanism was first introduced in the form of the pointer network \citep{vinyals2015pointer}. Expanding their work, various approaches combining attention probability and generation probability have been proposed \citep[]{miao-blunsom-2016-language, gu-etal-2016-incorporating, gulcehre-etal-2016-pointing, nallapati-etal-2016-abstractive}. In particular, \citet{see-etal-2017-get} proposed PGNet, where two distributions are weighted and merged into a single mixture distribution with the aid of a soft switch mechanism. PGNet has become \emph{de facto standard} for copy mechanisms in various tasks, including summarization \citep[]{paulus2018deep, gehrmann-etal-2018-bottom}, data-to-text \citep[]{gehrmann-etal-2018-end, rebuffel2020hierarchical}, and question answering \citep{McCann2018decaNLP}.

Table \ref{table1-copymodels} shows the characteristics of the copy models.
While our models bear a resemblance to models that trains the switch probability, ours are considerably different from theirs in two aspects: (i) Compared to \citet[]{gulcehre-etal-2016-pointing, nallapati-etal-2016-abstractive, chen-etal-2020-shot, wu-etal-2020-question}, those works train their copy components to activate only for certain words (i.e., OOV words, named entites, keywords, or values of structured data) whereas we activate it for all copy-candidates. Our \emph{force-copy-unk} model also seems like it trains to copy only for unknown words, but it works for all copy-candidates by maintaining attention loss (see Eq. \ref{loss_attn}).
(ii) We retain the vocabulary loss (see Eq. \ref{loss_vocab}) even if the model copies a word whereas the other models induce a trade-off between copy and generation. Models that has a trade-off between copy and generation definitely weaken the effect of vocabulary loss when they activate the copy process. However, we believe it's necessary to keep vocabulary loss to prevent mistaken prediction from the vocabulary distribution because the switch probability \(p_{gen}\) takes value in scalar range of \([0, 1]\), not a binary value.

\section{Experiments}
\label{sec:experiments}
\subsection{Data-to-text Generation}
\textbf{Experimental Settings}
\\
We perform experiments on using two datasets, i.e., \rotowire \citep{wiseman-etal-2017-challenges} and MLB \citep{puduppully-etal-2019-data} for data-to-text generation tasks. \rotowire contains professionally written articles summarizing NBA basketball games paired with corresponding game statistics, whereas MLB contains summaries paired with MLB baseball games. Both datasets consist of table-structured data records and relatively long multi-sentence documents (337 and 542 words on average for \rotowire and MLB, respectively). Note that MLB dataset we used to train and test our model slightly differs from Puduppully et al. in two aspects.\footnote{In table \ref{table2-d2tresult}, CS and CO scores of the MLB GOLD test set are under 100. This result indicates the difference of dataset between ours and \citet{puduppully-etal-2019-data}}
First, part of the official data has been modified since Puduppully et al. Second, we modified the script to preprocess the summaries to adjust to our model. Hence, we only compare the models that were trained on our version of the dataset. 
We compare our model against (i) \textbf{Templ}ate-based generators from \citet{wiseman-etal-2017-challenges} for \rotowire and from \citet{puduppully-etal-2019-data} for MLB; (ii) \textbf{WS-2017}, a standard encoder-decoder system with copy mechanism \citep{wiseman-etal-2017-challenges}; (iii) \textbf{RBF-2020}, a Transformer architecture with a hierarchical attention mechanism over entities and records within entities \citep{rebuffel2020hierarchical} based on \emph{OpenNMT + force copy attn} \citep{klein-etal-2020-opennmt} (See Table \ref{table1-copymodels}); (iv) \textbf{ENT}, the entity-based model of \citet{puduppully-etal-2019-data} that creates dynamically updated entity-specific representations; (v) \textbf{PGNet}, a Pointer-Generator Network \citep{see-etal-2017-get} which we re-implemented based on the hierarchical-attention architecture of RBF-2020. Note that our (\textbf{FC}: \emph{force-copy}, \textbf{FCU}: \emph{force-copy-unk}) models are also based on RBF-2020, in which we modified the copy module from the original. More details can be found in the Appendix A.1.
\vspace{5pt}\\
\textbf{Results}
\begin{table}[]
\small
\centering
\begin{tabular}{@{}L{1.57cm}@{}|R{0.57cm}R{0.59cm}|R{0.57cm}R{0.59cm}|R{0.64cm}|@{\hspace{3pt}}R{0.8cm}@{ }}
\toprule
\multicolumn{1}{c|}{\multirow{2}{*}{\textbf{RW}}} & \multicolumn{2}{c|}{\textbf{RG}} & \multicolumn{2}{c|}{\textbf{CS}} & \multirow{2}{*}{\textbf{CO}} & \multirow{2}{*}{\textbf{BLEU}} \\
\multicolumn{1}{c|}{} & \textbf{P\%} & \textbf{\#} & \textbf{P\%} & \textbf{R\%} &  &  \\ \midrule
GOLD & 96.11 & 17.31 & \textcolor{white}{.\_}100 & \textcolor{white}{.\_}100 & \textcolor{white}{.\_}100 & 100 \\
Templ & 99.95 & 54.15 & 23.74 & 72.36 & 11.68 & 8.9 \\
WS-2017 & 75.62 & 16.83 & 32.80 & 39.93 & 15.62 & 14.2 \\
RBF-2020 & 89.46 & 21.17 & 39.47 & 51.64 & 18.90 & 17.5 \\
ENT & 92.69 & 30.11 & 38.64 & 48.50 & 20.17 & 16.2 \\ \midrule
PGNet & 87.14 & 20.98 & \textbf{40.53} & 48.06 & \textbf{19.80} & 16.3 \\
RBF-2020\textsuperscript{†} & 89.31 & 22.07 & 36.88 & \textbf{49.37} & 17.87 & \textbf{16.6} \\
Ours(FC) & 93.27 & 24.28 & 34.34 & 48.85 & 17.26 & 15.8 \\
Ours(FCU) & \textbf{95.40} & \textbf{27.37} & 30.65 & 48.39 & 15.14 & 14.2 \\ \bottomrule \addlinespace[\belowrulesep]
\end{tabular}
\begin{tabular}{@{}L{1.57cm}@{}|R{0.57cm}R{0.59cm}|R{0.57cm}R{0.59cm}|R{0.64cm}|@{\hspace{3pt}}R{0.8cm}@{ }}
\toprule \addlinespace[\aboverulesep]
\multicolumn{1}{c|}{\multirow{2}{*}{\textbf{MLB}}} & \multicolumn{2}{c|}{\textbf{RG}} & \multicolumn{2}{c|}{\textbf{CS}} & \multirow{2}{*}{\textbf{CO}} & \multirow{2}{*}{\textbf{BLEU}} \\
\multicolumn{1}{c|}{} & \textbf{P\%} & \textbf{\#} & \textbf{P\%} & \textbf{R\%} &  &  \\ \midrule
GOLD & 92.07 & 21.02 & 98.84 & 99.91 & 98.76 & 100 \\
Templ & 97.96 & 59.93 & 68.46 & 22.82 & 10.64 & 3.8 \\ \midrule
PGNet & 79.12 & 20.94 & 47.06 & 47.73 & 20.07 & 9.9 \\
RBF-2020\textsuperscript{†} & 81.71 & 19.53 & 47.75 & 47.11 & 19.37 & 9.7 \\
Ours(FC) & 82.50 & \textbf{21.71} & 46.91 & 49.01 & 19.28 & 10.4 \\
Ours(FCU) & \textbf{84.50} & 21.05 & \textbf{49.39} & \textbf{50.89} & \textbf{21.16} & \textbf{10.5} \\ \bottomrule
\end{tabular}
\caption{\label{citation-guide}
Evaluation on the \rotowire and MLB test sets using RG count (\#) and precision (P\%), CS precision (P\%) and recall (R\%), CO in normalized Damerau-Levenshtein distance (DLD\%), and BLEU. The bottom section corresponds to our implementation of the corresponding methods. † denotes the duplicate model.}
\label{table2-d2tresult}
\end{table}
\\Following prior work (\citealp{wiseman-etal-2017-challenges}; \citealp{puduppully2019data}, \citeyear{puduppully-etal-2019-data}; \citealp{rebuffel2020hierarchical}), we evaluate our models using both extractive (relation generation (RG), content selection (CS), and content ordering (CO)) and \(n\)-gram metrics (BLEU). For extractive metrics, we use the Information Extraction (IE) system suggested by \citet{wiseman-etal-2017-challenges}. Given a data record \(r\), gold summary \(y\) and generated text \(\hat{y}\), the IE system identifies the entity (e.g., Knicks) and value (e.g., 28) pairs from \(\hat{y}\) and then predicts the pair relation (e.g., WIN). RG measures precision and the number of unique relations extracted from \(\hat{y}\) that also appear in \(r\). CS estimates precision and recall of the relations extracted from \(\hat{y}\) that are also extracted from \(y\). Finally, CO computes the normalized Damerau-Levenshtein distance (DLD) between the sequences of relations extracted from \(\hat{y}\) and \(y\). We used the pretrained IE model developed by \citet{wiseman-etal-2017-challenges} and \citet{puduppully-etal-2019-data} for \rotowire and MLB, respectively.

Table \ref{table2-d2tresult} shows our main results on the \rotowire and MLB datasets. Our \emph{force-copy} and \emph{force-copy-unk} models achieve higher RG precision for both corpora than any other neural models. Specifically, the RG precision of the \emph{force-copy-unk} model is as high as that of GOLD results for the \rotowire dataset. With respect to the loss function, the intervention of the copy mechanism increases in the order of PGNet, RBF-2020 and our models. We observe that RG precision performance increases along with this order. This effect is observed in both datasets.

The ENT system yields the highest scores in terms of CO on \rotowirestrip. Considering the model architectures, we conjecture that ENT leverages entity-specific representations as it specialized on data-to-text tasks, whereas the other neural models feature vanilla sequence-to-sequence structures.
\subsection{Abstractive Summarization}\label{experiment_abs}
\begin{table}[]
\centering
\small
\begin{tabular}{L{3.2cm}C{0.92cm}C{0.92cm}C{0.92cm}}
\toprule
\textbf{Method} & \textbf{R-1} & \textbf{R-2} & \textbf{R-L} \\ \midrule
Abstractive Model\textsuperscript{*} & 35.46 & 13.30 & 32.65 \\
ML + Intra-Attention & 38.30 & 14.81 & 35.49 \\
Pointer-Generator & 36.44 & 15.66 & 33.42 \\ \midrule
Pointer-Generator\textsuperscript{†} & 38.66 & 16.97 & 35.61 \\
Force-copy & 38.76 & 16.84 & 35.42 \\
Fore-copy-unk & \textbf{39.31} & \textbf{17.13} & \textbf{36.25} \\ \bottomrule
\end{tabular}
\caption{\label{citation-guide}
Evaluation on CNN/DailyMail test set using full-length ROUGE-F1 metric. The Bottom section corresponds to our implementation, and † denotes the duplicated model. * marked model used the anonymized dataset, so it is not strictly comparable to our results.}
\label{table3-rouge}
\end{table}
\begin{table}[]
\centering
\small
\begin{tabular}{L{3.4cm}C{3.4cm}}
\toprule
\textbf{Method} & \textbf{Copy Precision} \\ \midrule
Pointer-Generator & 47.80\% \\
Force-copy & 47.81\% \\
Force-copy-unk & \textbf{48.84\%} \\ \bottomrule
\end{tabular}
\caption{\label{citation-guide}
Evaluation on CNN/DailyMail test set using Copy Precision.}
\label{table4-copyprecision}
\end{table}
\textbf{Experimental Settings}\\
We use the non-anonymized version of the CNN/DailyMail dataset (\citealp{hermann-etal-2015-teaching}; \citealp{nallapati-etal-2016-abstractive}) provided by Harvard NLP and conform all experimental conditions presented by \citet{see-etal-2017-get}. More details can be found in the Appendix A.2.

Since the experimental purpose is not recording the highest ROUGE score but validating the performance of copy mechanism itself, we confine the experiments to encoder-decoder abstractive baselines trained with cross-entropy and do not adopt any additional techniques such as coverage or selector mechanisms. For that same reason, we re-implemented the Pointer-Generator network to exquisitely compare the effect of copy mechanism. 
Therefore, we compare our model against (i) \textbf{Abstractive model} \citep{nallapati-etal-2016-abstractive}, a pointer-based encoder-decoder model using two softmax layers; (ii) \textbf{ML+Intra-Attention} \citep{paulus2018deep}, an intra-attention model based on the encoder-decoder network; and (iii) \textbf{Pointer-Generator} \citep{see-etal-2017-get} and our re-implemented version comprising the modification of the original LSTM encoder to a Transformer encoder.
\vspace{5pt}\\
\textbf{Automatic Evaluation Results}\\
We adopt the ROUGE \citep{lin-2004-rouge} and Copy Precision (CP) for the evaluation metric. Given a source article \(x\), gold summary \(y\), and generated text \(\hat{y}\), CP estimates how well \(\hat{y}\) are matched to \(y\) for those words that appear in \(x\). 

Table \ref{table3-rouge}, \ref{table4-copyprecision} shows our ROUGE and CP evaluation results on the CNN/DailyMail corpus. The results of the baseline and the \emph{force-copy} model are close, where the \emph{force-copy} is better at R-1 and the baseline is better at R-2 and R-L. The \emph{force-copy-unk} model outperforms the baseline in all ROUGE criteria. This is consistent with the fact that \emph{force-copy-unk} outperforms on CP than the baseline. %

To investigate the copy and generation ratio, we report the novel \(n\)-gram scores in Table \ref{table5-novelngram} for the ground-truth and model-generated summaries, where it represents the level of abstractness \citep{kryscinski-etal-2018-improving}. Our \emph{force-copy} model produces less novel expression than the baseline. As the final generation probability is calculated by mixing the probabilities from the vocabulary and copy distributions, it is obscure to exactly determine that a word is copied or generated. However, we conjecture that \emph{force-copy} model tends to favor copying since 83\% of the target words are copy-candidates in the dataset that are trained to be copied. At inference time, it becomes more skewed owing to unprovided word-by-word supervision. Practically, the average \(p_{copy}\) (equivalent to \(1-p_{gen}\)) of the \emph{force-copy} model for all words of test set is 0.848, whereas that of the Pointer-Generator is 0.773. Contrarily, the \emph{force-copy-unk} model with an average \(p_{copy}\) of 0.018 tends to favor generation. This effect makes the \emph{force-copy-unk} model produce significantly more novel expressions than the baseline. In fact, \citet{kryscinski-etal-2018-improving} reported that there exists an inverse correlation between the ROUGE and novelty scores in all model types. Hence, the main benefit of our \emph{force-copy-unk} model is the increase in novel expression, as the model enhanced the ROUGE score simultaneously.
\begin{table}[]
\small
\begin{tabular}{@{}L{2.88cm}R{0.73cm}R{0.73cm}R{0.73cm}R{0.73cm}}
\toprule
\textbf{Method} & \textbf{NN-1} & \textbf{NN-2} & \textbf{NN-3} & \textbf{NN-4} \\ \bottomrule \addlinespace[\belowrulesep]
\emph{Ground truth Summary} & 16.99 & 56.19 & 73.98 & 82.38 \\ \midrule
Pointer-Generator & 0.25 & 5.37 & 11.45 & 16.85 \\
Force-copy & 0.03 & 3.54 & 8.10 & 12.36 \\
Force-copy-unk & \textbf{0.28} & \textbf{6.62} & \textbf{13.60} & \textbf{19.54} \\ \bottomrule
\end{tabular}
\caption{\label{citation-guide}
Comparison of novel \(n\)-gram (NN-) test results for our model and baselines on the CNN/DailyMail test set.}
\label{table5-novelngram}
\end{table}
\begin{table}[]
\centering
\small
\begin{tabular}{@{}L{1.6cm}|R{2.05cm}|R{0.7cm}R{0.7cm}R{0.7cm}}
\toprule
 & \textbf{Comparison} & \multicolumn{3}{r}{\textbf{Preference(\%)}} \\
 & \textbf{(A vs B)} & \textbf{A} & \textbf{B} & \textbf{Tie} \\ \midrule
\multirow{3}{*}{Readability} & PGNet vs. FC & 6.0 & 12.0 & 82.0 \\
 & PGNet vs. FCU & 10.0 & 8.7 & 81.3 \\
 & FC vs. FCU & 12.6 & 6.7 & 80.7 \\ \midrule
\multirow{3}{*}{Factuality} & PGNet vs. FC & 4.7 & 16.0 & 79.3 \\
 & PGNet vs. FCU & 20.7 & 11.3 & 68.0 \\
 & FC vs. FCU & 28.7 & 2.0 & 69.3 \\ \midrule
\multirow{3}{*}{Abstractness} & PGNet vs. FC & 27.3 & 9.3 & 63.3 \\
 & PGNet vs. FCU & 12.7 & 20.7 & 66.7 \\
 & FC vs. FCU & 4.0 & 31.3 & 64.7 \\ \bottomrule
\end{tabular}
\caption{\label{citation-guide}
Human evaluation results on the 50 randomly sampled articles of CNN/DailyMail test sets. Each summary pair is reviewed by 3 human evaluators. Agreement scores by Fleiss’ kappa \citep{fleiss1971mns} are 0.19 for Readability, 0.44 for Factuality and 0.29 for Abstractness, respectively.}
\label{table6-humaneval}
\end{table}
\vspace{5pt}\\
\textbf{Human Evaluation Results}
\\We conduct a human evaluation to measure the quality of the summaries of each model. The instructions are related to three different aspects, as follows. (i) Readability: How well-written (fluent and grammatical) is the summary? (ii) Factuality: Is the summary factually consistent with the source document? (iii) Abstractness: How does the summary generate novel expressions (not entirely copy the source sentences)?

As shown in Table \ref{table6-humaneval}, our \emph{force-copy} model outperforms the others on readability and factuality and \emph{force-copy-unk} model outperforms the others on abstractness. It seems that more copy operations make the summary be good at readability and factuality levels but results in poor abstractness. We discuss this further in Section \ref{sec:trade-off}.
\begin{table}[]
\small
\centering
\begin{tabular}{@{}L{2.4cm}R{2.2cm}R{2.2cm}@{}}
\toprule
\multirow{2}{*}{\textbf{Method}} & \textbf{RG(P)} & \textbf{RG(P)} \\
 & \textbf{\(p_{copy} > 0.5\)} & \textbf{\(p_{gen} >0.5\)} \\ \midrule
Pointer-Generator & 0.8043 (57.1\%) & 0.7757 (42.9\%) \\
Force-copy & \textbf{0.8250} (62.3\%) & 0.7797 (37.7\%) \\
Force-copy-unk & - (0\%) & \textbf{0.8365} { }(100\%) \\ \bottomrule
\end{tabular}
\caption{\label{citation-guide}
Comparison of RG Precision (for only unique and numeric values of \rotowirestrip) for our model and baseline depending on copy probability. Percentage numbers represent the ratio of the number of words that \(p_{copy}>0.5\) and \(p_{gen}>0.5\).}
\label{table7-probandrg}
\end{table}
\section{Analysis and Discussion}
\subsection{Effects of Copy Probability on Precision}\label{sec:effect-of-cp}
To examine the effects of copy probability on generation precision, we investigate the word-level \(p_{copy}\) on the \rotowire test set. Based on the IE system, we report the RG precision depending on the copy probability on Table \ref{table7-probandrg}. To couple the IE system and generated summary, we limit this study only to unique and numeric words. Hence, this RG precision is different from that shown in Table \ref{table2-d2tresult}. 

We first observe that high copy probability results in high precision in both \emph{force-copy} model and the baseline, which adheres to the goals of the copy mechanism. Although the precision gap between two models for \(p_{gen}>0.5\) is only 0.4\%p, it increases to 2.1\%p for \(p_{copy}>0.5\). These findings confirm that our \emph{force-copy} model shows high confidence on copy probability than the baseline. Furthermore, it generates 62.3\% of \(p_{copy}>0.5\) words compared to 57.1\% of the baseline, further widening the precision gap. Meanwhile, the \emph{force-copy-unk} model does not generate \(p_{copy}>0.5\) words for the values of the \rotowire test set as the numeric words exist in the target vocabulary. Although the \emph{force-copy-unk} model results in low \(p_{copy}\), attention loss (Eq. \ref{loss_attn}) prompts the generation to look up the values to copy during training. This copy-like generation corresponds to 6.1\%p precision increase compared to the baseline for the words \(p_{gen}>0.5\).
\subsection{Impact of Vocabulary Size on the Force-copy-unk Model}
As copy mechanism of our \emph{force-copy-unk} model relies on a target vocabulary, we study the effects of vocabulary size on the \emph{force-copy-unk} model. In Section \ref{experiment_abs}, we experimented on the abstractive summarization task on the CNN/DailyMail corpus with 50k vocabularies. The vocabulary coverage is 72.1\%, and 1.9\% words of the gold test set summaries are OOV. When we reduce the vocabulary size to 25k, the  percentages become 53.3\% and 3.5\%, respectively. In contrast when we increase the vocabulary size to 100k, the percentages become 83.6\% and 1.1\%, respectively.
Table \ref{table8-vocabcomp} shows a relative comparison in terms of vocabulary size. As reported by \citet{see-etal-2017-get}, large vocabulary size does not enhance the performance of attention-based sequence-to-sequence models. This is consistent with our observations that large (100k) vocabulary models perform poorly in terms of ROUGE score. Even 50k-vocabulary model falls behind the 25k model on ROUGE score in practice. In contrast, we observe novel \(n\)-gram expression increases as vocabulary size grows. This disparity is likely associated with less vocabulary size during training, which prompts the model to copy more (see Eq. \ref{eq:force-copy-unk}). According to the above results, the \emph{force-copy-unk} model can control this phenomenon by adjusting the vocabulary size. %
\subsection{Tradeoff between Copy and Abstractness}\label{sec:trade-off}
\begin{table}[]
\centering
\small
\begin{tabular}{@{}L{0.65cm}|@{ }R{0.6cm}R{0.6cm}R{0.6cm}|@{}R{0.91cm}@{ }R{0.91cm}@{ }R{0.91cm}@{ }R{0.91cm}@{}}
\toprule
\textbf{\begin{tabular}[c]{@{}l@{}}\(\|V\|\)\end{tabular}} & \textbf{R-1} & \textbf{R-2} & \textbf{R-L} & \textbf{NN-1} & \textbf{NN-2} & \textbf{NN-3} & \textbf{NN-4} \\ \midrule
25k & \textbf{39.31} & \textbf{17.13} & \textbf{36.25} & 0.28 & 6.62 & 13.60 & 19.54 \\
50k & 38.52 & 16.62 & 35.58 & 0.46 & 7.96 & 15.85 & 22.42 \\
100k & 38.11 & 16.21 & 35.28 & \textbf{0.73} & \textbf{10.47} & \textbf{20.14} & \textbf{27.84} \\ \bottomrule
\end{tabular}
\caption{\label{citation-guide}
ROUGE and novel \(n\)-gram comparison in terms of vocabulary size for our force-copy-unk model.}
\label{table8-vocabcomp}
\end{table}
\begin{table}[]
\centering
\small
\setlength\extrarowheight{2pt}
\begin{tabular}{|M{7.3cm}|}
\hline
\textbf{Article} no other scoreline could have emphasised how seriously this new arsenal team are taking their prospects of winning major trophies again (…) olivier giroud wants arsenal to continue their premier league winning streak and challenge for the title. (…) arsenal to earn a 1-0 victory over burnley on saturday. (…) \\ \hline
\textbf{PGNet} arsenal beat burnley 1-0 in the premier league on saturday. \textcolor{red}{olivier giroud wants arsenal to continue their premier league winning streak and challenge for the title.} \\ \hline
\textbf{Force-copy} \textcolor{red}{no other scoreline could have emphasised how seriously this new arsenal team are taking their prospects of winning major trophies again. olivier giroud wants arsenal to continue their premier league winning streak and challenge for the title.} \\ \hline
\textbf{Force-copy-unk} arsenal beat burnley 1-0 in their premier league clash on saturday. olivier giroud feels the new arsenal team (…) \\ \hline
\end{tabular}
\captionof{figure}{\label{citation-guide}
Examples of generated summaries of CNN/DailyMail of each model. (\textcolor{red}{red} color denotes copied sentences from the source article)}
\label{figure3}
\end{table}
The human evaluation results presented in Section \ref{experiment_abs} indicate that our \emph{force-copy} model outperforms other models in terms of readability and factuality. However, the \emph{force-copy-unk} model exhibits low readability and factuality in abstractive summarization tasks. This is a conflicting outcome given that the \emph{force-copy-unk} model enhanced the RG precision of data-to-text tasks. It likely attributed to the difference of two tasks and datasets. As shown in Figure \ref{figure1}, only values of structured data (i.e., names and records) can be copied in data-to-text task. In contrast, almost every word can be copied in summarization task. This is consistent with \citet{song-etal-2020-controlling}, which found that the proportion of copy-candidate words used for training impacts the model to depend on copying the words, leading to pure extracts and less abstraction. The examples in Figure \ref{figure3} show that the more the model activates copy during training, the more it behaves like an extractive summarization model. Abstraction introduces even more choice of phrasing, negatively affecting factuality and readability compared to the extracted source sentence. In any case, encouraging the model to write more abstractly while retaining the factuality and readability is an interesting subject for the future work.

\section{Conclusion}
In this work we analyzed the prior copy models in detail. Based on the analysis, we presented a novel copy mechanisms that leverages Pointer-Generator network. We showed that our models conduct more accurate copy than baselines via data-to-text experiment. In addition, our \emph{force-copy-unk} model outperformed the baselines in both ROUGE score and the novel \(n\)-gram score in abstractive summarization task.

\bibliography{anthology,custom}
\bibliographystyle{acl_natbib}

\clearpage
\input{Appendix}

\end{document}

%% file: Appendix.tex
\pdfoutput=1

\newcolumntype{M}[1]{>{\arraybackslash}p{#1}}
\newcolumntype{L}[1]{>{\raggedright\arraybackslash}p{#1}}
\newcolumntype{C}[1]{>{\centering\arraybackslash}p{#1}}
\newcolumntype{R}[1]{>{\raggedleft\arraybackslash}p{#1}}
\definecolor{theblue}{RGB}{0, 112, 192}
\definecolor{thegreen}{RGB}{44, 150, 72}
\definecolor{theorange}{RGB}{242, 150, 51}

\appendix

\section{Experimental Details}
\label{sec:exp-details}
\subsection{Data-to-text Generation}\label{appendix:data-to-text}
We used the official train/valid/test splits of both datasets for our experiments (3,398/727/728 for \rotowire and 22,821/1,739/1,744 for MLB).\footnote{The complete ROTOWIRE datasets that we used can be obtained from \url{https://github.com/harvardnlp/boxscore-data}, and MLB datasets from \url{https://github.com/ratishsp/mlb-data-scripts}.}

For \rotowirestrip, We adopt word embedding size of 300, and feature embedding size of 27. We employed a Transformer encoder based on the hierarchical architecture of RBF-2020. The encoder consists of 3 low-level (unit) layers and 3 high-level (chunk) layers, where each layer comprises feed-forward layer with 1024-dimensional hidden states. For decoder, We adopt 2-layer LSTM network. The number of parameters of the model in this setting is 14,152,201. We additionally tried a bigger embedding size (600 and 100 for word embedding and feature embedding by max, respectively) and bigger encoder layer size (6 by max), resulting in inconsistent or ineffective outcomes in terms of RG precision. We train the model using Adam with learning rate of 1e-3 and weight decay value of 1e-5. Training took 13 hours for 18k training iterations on a single V100 GPU with a batch size of 96. During inference, we use beam search with beam size of 10 and block \(n\)-gram repeat over 10 words. Every experiment has been conducted at least twice, and we report the best results among them.

For MLB, we adopt word embedding size of 512, and feature embedding size of 32. The encoder consists of 4 low-level layers and 4 high-level layers, where each layer comprises feed-forward layer with 1024-dimensional hidden states. The number of parameters of the model in this setting is 69,562,999. Training took 7 days and 8 hours for 42k training iterations on a single V100 GPU with a batch size of 96. Other settings are the same as those of the \rotowirestrip.

\subsection{Abstractive Summarization}\label{appendix:abs}
We used the 287,227/13,368/11,490 train/valid/test splits of CNN/DailyMail datasets for our experiments, which is provided by Harvard NLP.\footnote{the official splits of CNN/DailyMail datasets can be obtained from \url{https://github.com/harvardnlp/sent-summary}.}
Additionally, we truncate the article to 400 tokens and control the length of the summary to less than 100 tokens for training and less than 120 tokens for inference. 
According to the original paper, we use 50K vocabulary for the Pointer-Generator, where the author noted the best result. We applied 50K, 25K vocabulary size for the \emph{force-copy}, \emph{force-copy-unk} respectively, for the best performance. We include more detailed analysis about the effect of vocabulary size change in Section 6.2 of the main paper.

We use word embedding size of 512 and adopt a 6-layer Transformer encoder and a 3-layer LSTM decoder without any pre-trained word embeddings. Note that we shared the source and target embeddings since the same vocabularies are used in the the source and the target data in this task. The number of parameters of the model in this setting is 51,136,213. We train using Adam with learning rate of 2e-4 and weight decay value of 1e-5. We train our models and baseline models for approximately 40 epochs. Training took 2 days and 15 hours on a single V100 GPU with a batch size of 96. During decoding, we implement beam search with beam size of 10, and block \(n\)-gram repeat over 3 words.

For the evaulation metric, we used official ROUGE-1.5.5 perl script (\texttt{ROUGE-1.5.5.pl}) provided by a python wrapper package. \footnote{\url{https://github.com/falcondai/pyrouge/}}

\newpage
\onecolumn
\section{Examples}\label{appendix:exp-detail}
\subsection{Data-to-text Generation}
\begin{table}[hbt!]
\small
\centering
\begin{tabular}{C{2.7cm}C{0.8cm}C{0.8cm}C{0.8cm}C{0.7cm}C{1.1cm}C{1.1cm}C{0.75cm}C{1.2cm}C{1.2cm}C{0.3cm}}%
\toprule
TEAM            & HOME & WIN & LOSS & PTS & FG\_PCT & FG3\_PCT & RB & PTS\_QTR1 & PTS\_QTR4 & ... \\ \midrule
Boston Celtics  & YES & 21  & 19   & 117 & 44      & 47       & 56 & 36       & 21       & ... \\
Pheonix Suns    & NO & 13  & 28   & 103 & 39      & 37       & 47 & 29       & 25       & ... \\ \addlinespace[\aboverulesep] \toprule 
\end{tabular}
\begin{tabular}{C{3cm}C{0.58cm}C{0.58cm}C{0.58cm}C{0.58cm}C{0.58cm}C{0.6cm}C{0.6cm}C{0.58cm}C{0.58cm}C{0.58cm}C{0.58cm}C{0.58cm}C{0.2cm}}
PLAYER          & AS  & RB  & PT  & FGM  & FGA & FG3M & FG3A & FTM & FTA & BLK & STL & MIN & ... \\ \midrule
Isaiah Thomas & 5 & 2 & 19 & 5 & 11 & 3 & 5 & 6 & 6 & 0 & 0 & 24 & ...\\
Sonny Weems & 5 & 6 & 10 & 4 & 6 & 1 & 1 & 1 & 2 & 2 & 1 & 13 & ...\\
Jarad Sullinger & 2 & 10 & 11 & 4 & 8 & 1 & 2 & 2 & 4 & 0 & 0 & 24 & ...\\ \bottomrule
\end{tabular}
\begin{tabular}{c}
\\
\end{tabular}
\normalsize
\setlength\extrarowheight{2pt}
\begin{tabular}{|M{15.3cm}|}
\hline
\textbf{Pointer-Generator} \textcolor{theblue}{The Boston Celtics ( 21 - 19 ) defeated} \textcolor{red}{the host Sonny Weems Weems Sonny ( 4 - 6 )} , \textcolor{theblue}{117} - \textcolor{red}{6} , \textcolor{theorange}{in a nail - biter on Saturday} . This was a tight game throughout the first half , as \textcolor{red}{the Celtics outscored the Sixers 36 - 6} \textcolor{theorange}{in the final 12 minutes} . \textcolor{theblue}{The Celtics were the superior shooters , as they shot 44 percent from the field} and \textcolor{red}{47 percent from the free - throw line} and \textcolor{theblue}{47 percent from three - point range} . The rebounding was a huge factor as well , \textcolor{theblue}{with Boston winning that battle , 56} - \textcolor{red}{6} . \textcolor{theblue}{The Celtics also dominated the rebounding , winning that battle} , \textcolor{theblue}{56} - \textcolor{red}{6} . (…)\\ \hline
\textbf{RBF-2020} \textcolor{theblue}{The Boston Celtics ( 21 - 19 ) defeated the Phoenix Suns ( 13 - 28 ) 117 - 103} \textcolor{theorange}{on Sunday} \textcolor{thegreen}{at the TD Garden in Boston} . The Celtics got off to a quick start in this one , \textcolor{theblue}{out - scoring the Suns 36 - 29 in the first quarter} . \textcolor{red}{The Celtics were able to out - score the Suns 21 - 25 in the fourth quarter to secure the victory in front of their home crowd} . \textcolor{theblue}{The Celtics were the superior shooters in this one , going 44 percent from the field and 47 percent from the three - point line , while the Suns went 39 percent from the floor and a meager 37 percent from beyond the arc} . (…)\\\hline
\textbf{Force-copy} \textcolor{theblue}{The Boston Celtics ( 21 - 19 ) defeated the Phoenix Suns ( 13 - 28 ) 117 - 103} \textcolor{theorange}{on Wednesday} \textcolor{thegreen}{at the US Airways Center in Phoenix} . The Celtics were the superior shooters in this game , \textcolor{theblue}{going 44 percent from the field and 47 percent from the three - point line , while the Suns went 39 percent from the floor and a meager 37 percent from beyond the arc} . The Celtics were led by \textcolor{red}{the duo of} \textcolor{theblue}{Isaiah Thomas , who went 5 - for - 11 from the field and 3 - for - 5 from the three - point line to score 19 points , while also adding five assists} . \textcolor{theorange}{It was his second} \textcolor{red}{double - double} \textcolor{theorange}{in a row} (…) \\\hline
\textbf{Force-copy-unk} \textcolor{theblue}{The Boston Celtics ( 21 - 19 ) defeated the Phoenix Suns ( 13 - 28 ) 117 - 103} \textcolor{theorange}{on Wednesday} \textcolor{thegreen}{at the TD Garden in Phoenix} . The Celtics got off to a quick start in this one , \textcolor{theblue}{out - scoring the Suns 36 - 29 in the first quarter alone} . \textcolor{theblue}{The Celtics were the superior shooters in this one , going 44 percent from the field and 47 percent from the three - point line , while the Suns went just 39 percent from the floor and 37 percent from beyond the arc} . (…) \textcolor{theblue}{Jared Sullinger} was \textcolor{red}{the only other starter} \textcolor{thegreen}{to reach double figures in points} , \textcolor{theblue}{as he finished with 11 points ( 4 - 8 FG , 1 - 2 3Pt , 2 - 4 FT ) and 10 rebounds in 24 minutes} .(…)\\\hline

\end{tabular}
\captionof{figure}{Comparison of output of data-to-text models on a \rotowire test set. We highlight text in \textcolor{theblue}{blue} if it accurately reflects a record, in \textcolor{red}{red} if it is inconsistent with the records, in \textcolor{thegreen}{green} if it can be inferred indirectly from the records (e.g., "at the TD Garden in Boston" can be inferred from the HOME column in the Boston Celtics row), and in \textcolor{theorange}{orange} if there are no conflicting or supporting records at all (even in the train set). Best viewed in color.}
\label{figure1}
\end{table}

\newpage
\onecolumn
\subsection{Abstractive Summarization}
\begin{table}[hbt!]
\normalsize
\setlength\extrarowheight{2pt}
\begin{tabular}{|M{15.3cm}|}
\hline
\textbf{Article (truncated)}\\
concerns are raised about labour 's policy under shadow education secretary tristram hunt . the heads of some of britain 's best state schools today warn of the dangers of a labour government reversing radical education reforms . in a letter to the daily mail , 80 current and former leaders say there is clear evidence that academy-style freedoms are benefiting a generation of children . but they say labour -- and some senior lib dems -- appear to be threatening to reimpose state controls . the letter , signed by the heads of good and outstanding autonomous schools , was backed yesterday by david cameron . in it , they claim there is evidence that the most successful education systems benefit from schools with academy-style freedoms . they say such schools are more likely to be ranked ` outstanding ' by ofsted and more likely to improve . ` secondary schools which have converted to academy status outperform other schools -- by a margin of almost 10 per cent , ' they wrote . but the heads expressed alarm at comments by ed miliband that labour would reimpose ` a proper local authority framework for all schools ' . senior lib dems were also accused of suggesting they no longer support freedom for acdemies , which are able to control pay , conditions and the curriculum . ` this is not the time to stop something that is working to the benefit of so many children in schools , ' wrote the heads . schools on the letter include torquay boys ' grammar school , ranked in the top 100 for gcse results this year . (...)\\
\hline
\textbf{Pointer-Generator}\\80 current and former leaders say there is clear evidence that academy-style freedoms are benefiting a generation of children .\\
but they say labour and some senior lib dems appear to be threatening to reimpose state controls .\\
\hline
\textbf{Force-copy}\\80 current and former leaders say there is clear evidence academy-style freedoms are benefiting a generation of children .\\
senior lib dems appear to be threatening to reimpose state controls .\\
\hline
\textbf{Force-copy-unk}\\shadow education secretary tristram hunt said labour 's policy is ` not the time to stop something that is working to the benefit of so many children in schools ' \\
but the heads of good and outstanding autonomous schools have converted to academy status outperform other schools -- by a margin of almost 10 per cent .\\
\hline

\end{tabular}
\captionof{figure}{Comparison of output of abstractive generation models on CNN/DailyMail test sets.} 
\end{table}

\newpage
\onecolumn
\begin{table}[hbt!]
\normalsize
\setlength\extrarowheight{2pt}
\begin{tabular}{|M{15.3cm}|}
\hline
\textbf{Article (truncated)}\\
it 's t20 season on the sub-continent and the world 's best players are about the pad up for the latest edition of the indian premier league , cricket 's most exciting and richest domestic tournament . eight teams will play a total of 60 games over almost seven weeks across 12 venues all round india in a battle to be crowned champions of the tournament 's eighth edition , with the final taking place at eden gardens in kolkata on may 24 . can kolkata knight riders retain their title ? will virat kohli lead the royal challengers bangalore to their first title ? can ms dhoni 's chennai super kings win their third crown ? and who are the players to watch out for ? sportsmail tells you all you need to know in our guide to the 2015 indian premier league as india prepares for the spectacular cricket roadshow . ms dhoni , pictured in the 2011 champions league , is looking to guide chennai super kings to a third title . chennai super kings . the bright yellow jerseys of the super kings are one of the iconic sights of the indian premier league . led by the superstar indian duo of ms dhoni and suresh raina , chennai are the most successful team in ipl history . as well as their back-to-back victories in 2010 and 2011 , csk have been losing finalists three times and never failed to reach the last four . in international players dhoni , raina , ravi ashwin , ravi jadeja and mohit sharma , the super kings have probably the best pool of indian talent in the tournament , which is key given that seven of the starting xi have to be domestic players . the foreign talent is also strong , though , and includes new zealand captain brendon mccullum , south african faf du plessis and west indian all-rounder dwyane bravo . one to watch : there are so many . dhoni needs no introduction , raina is the top scorer in ipl history , but mccullum is one of the most exciting players in world cricket at the moment  (...)\\
\hline
\textbf{Pointer-Generator}\\india face india in the indian premier league on may 24.\\
can kolkata knight riders retain their first title of the tournament .\\
can ms dhoni 's chennai super kings win their third crown ?\\
\hline
\textbf{Force-copy}\\ eight teams will play a total of 60 games across 12 venues all round india in a battle to be crowned champions of the tournament 's eighth edition , with the final taking place at eden gardens in kolkata on may 24 . \\
sportsmail tells you all you need to know in our guide to the 2015 indian premier league as india prepares for the spectacular cricket roadshow .\\
senior lib dems appear to be threatening to reimpose state controls .\\
\hline
\textbf{Force-copy-unk}\\ eight teams will play a total of 60 games over almost seven weeks across 12 venues all round india in a battle to be crowned champions of the tournament 's eighth edition .\\
chennai super kings are one of the iconic sights of the indian premier league .\\
brendon mccullum is one of the most exciting players in world cricket at the moment .\\
\hline

\end{tabular}
\captionof{figure}{Comparison of output of abstractive generation models on CNN/DailyMail test sets.} 
\end{table}